# Determining the Mechanical Properties of a New Composite Material using Artificial Neural Networks

Emilia Ciupan[1], Mihai Ciupan[2], Daniela-Corina Jucan[3]

[1]*Associate Professor, Department of Management and Economic Engineering, Technical University of Cluj-Napoca, Romania*

[2]*PhD. Student, Department of Manufacturing Engineering, Technical University of Cluj-Napoca, Romania*

[3]*Lecturer, Department of Management and Economic Engineering, Technical University of Cluj-Napoca, Romania*

*Abstract*

*The paper studies the possibility of using artificial neural networks (ANN) to determine certain mechanical properties of a new composite material. This new material is obtained by a mixture of hemp and polypropylene fibres. The material was developed for the industry of upholstered furniture. Specifically, it is intended for the making of elements of the support structure of some upholstered goods (chairs, armchairs, sofa sides) with the objective of replacing wood. The paper aims to calculate the following mechanical properties: maximum tensile strength and maximum elongation.*

## I. INTRODUCTION

Designing optimum products made of a new composite material requires good knowledge of the material's mechanical properties. These properties cannot be easily determined from its constituents. Furthermore, the composite's properties cannot be calculated from the constituents properties using mathematical formulas. Thus, determining them can only be done experimentally by using specialized machines and testing methods. Company Taparo SA from Târgu-Lăpuș, Romania, finds itself in this exact situation having developed a new composite material.

The material has two constituents: thermoplastic polypropylene fibres that are 20-60 mm long and linear mass density of 7-16 DEN, which make up 40-50% of the composite's mass, and hemp fibres that are 60-100 mm long and 70-80 denier linear mass density.

The material has been designed to replace wood in the making of structural elements of upholstered furniture (chairs, armchairs, sofas). Manufacturing these elements from the composite material previously presented implies redesigning them, but optimizing their shape using numerical simulation requires knowing the mechanical properties of the material. These include the tensile strength, elongation at break, Young's modulus and Poisson's ratio.

The composite material has a thickness that is determined by the weaving process, but multiple layers can be stacked together to obtain different thicknesses. After cutting the material according to the shape of the product (chair seat or backrest, armchair shell or sofa side), the multi-layered material is heated in an oven to a temperature of 200-220°C and is then thermoformed using a mould.

The multi-layered material's properties are influenced by the number of layers and their relative orientation to one another. This influence was investigated experimentally and was published in the paper [1].

Tests were conducted for different numbers of layers (2, 3 and 4 layers) and for different orientations.

We consider the material composed of 4 layers arranged as follows:

1. All the layers are oriented longitudinally (denoted LLLL);

2. Each layer is rotated 90° from the previous, so that we have alternating longitudinal and transverse layers (denoted LTLT).

Longitudinal refers here to the direction in which the material travels through the weaving process.

## II. PLANNING OF THE MEASUREMENT EXPERIMENT

The composite material made of 4 layers oriented as LLLL and LTLT and thermoformed as sheets was cut into test specimens with the following orientations: parallel to the longitudinal direction - L, diagonally - D (45° to the longitudinal direction) and transversely - T (90° to the longitudinal direction).

The tensile strength and elongation at break were determined on 25 mm wide specimens with 10 specimens being cut for each layer orientation (LLLL, LTLT) and cutting direction (L, D, T) according to standard D3039 "Test Method for Tensile Properties of Polymer Matrix Composite Materials" [2].

Figure 1 presents an example of specimens cut from an LLLL material with an L cutting direction.

The Zwick/Roell Z150 machine was used to perform the tensile tests. The following machine parameters were used [1]:
- 150 mm grip to grip separation at the start position [3];
- 0,4 MPa preload;
- 2 mm/min test speed.

The machine control, the data acquisition, as well as the processing and export of results were done using a PC and a program called testXpert II.

The graphical form by which testXpert II gives the results of the experiment consists of stress-strains graphs, as in Figure 2.

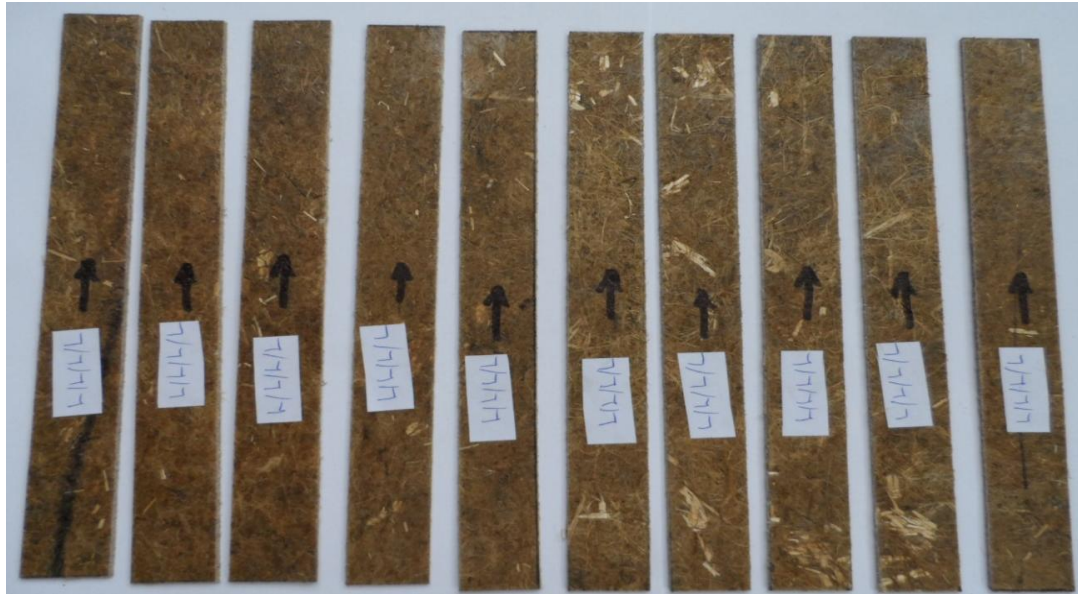

Figure 1. Specimens of type LLLL, cut on the L direction [2]

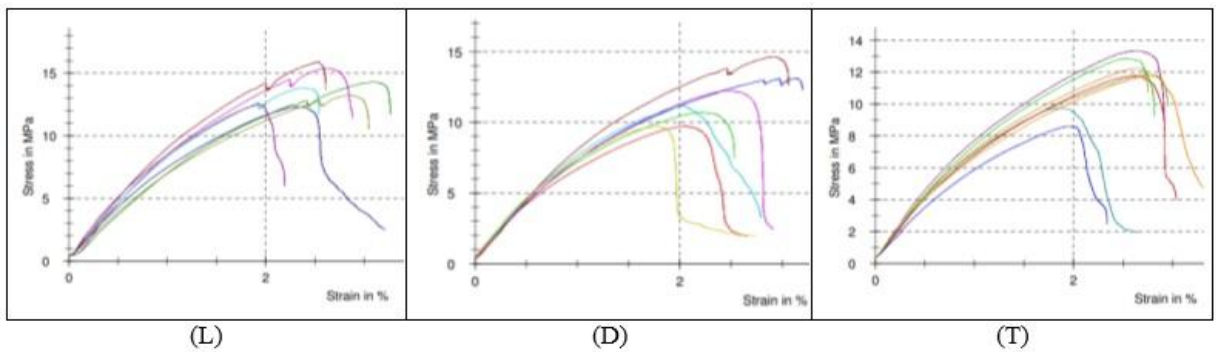

Figure 2. Stress – strain diagrams for LLLL-type specimens, cut on L, D and T directions

### III. SIMULATION OF THE MECHANICAL PROPERTIES OF THE MATERIAL USING ANN

There are numerous examples of using ANN for the successful modelling of the most diverse processes and activities. Activities that belong to the field of management [4][5] or different industrial processes [6][7] can be considered. Their applications are also known in the analysis and forecasting of weather conditions and climate change [8][9], with implications for agriculture [10]. More recently, questions arise about the possibilities of applying artificial intelligence to less common areas, such as legal field [11]. Some of these combine ANN with other intelligent control methods [12].

The paper's objective is to research the extent to which ANN can be used in determining the tensile strength and elongation at break of the previously discussed composite material. In order to do this a perceptron type ANN with 3 layers was designed (Figure 3):
- Inputs: layer orientation and cutting direction;
- Outputs: tensile strength $\sigma_M$ and  elongation at break $\varepsilon_M$;

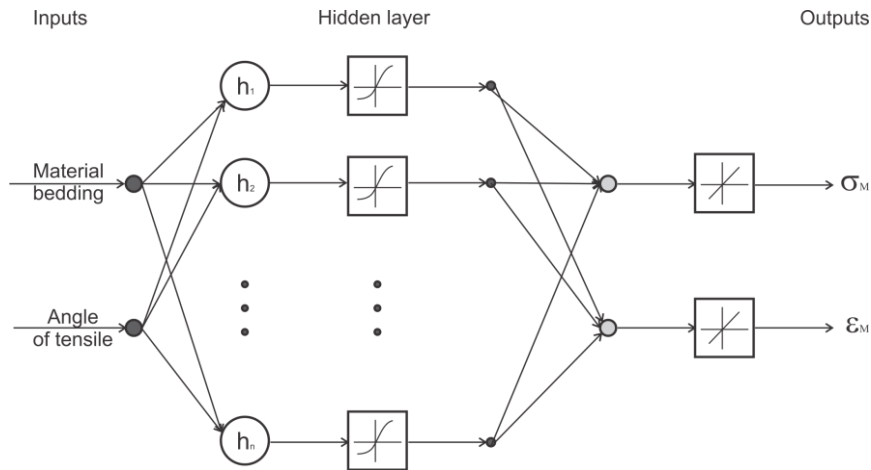

**Figure 3. Configuration of the ANN**

- Intermediate layer: a number of neurons chosen by trial and error according to the criterion of the minimum square root error obtained in the training of the network;
- The activation functions: the sigmoid function for the intermediate layer and the pureline function for the output layer;
- Training method: Levenberg-Marquardt.

*A. ANN Training*

The ANN training requires numerical values, so the two types of materials were coded as 1 for LLLL and 2 for LTLT.

The first half of the training data is partially presented in Table I.

**Table I**
**Training data**

| No. | Number of layers | Layout of the layers | Angle of the tensile [°] | $\sigma_M$ [Mpa] | $\varepsilon_M$ [%] | Mean values $\bar{\sigma}_M$ [Mpa] | $\bar{\varepsilon}_M$ [%] |
|---|---|---|---|---|---|---|---|
| (1) | (2) | (3) | (4) | (5) | (6) | (7) | (8) |
| 1 | 4 | 1 | 0 | 19.90 | 3.01 | | |
| 2 | 4 | 1 | 0 | 19.91 | 3.20 | | |
| 3 | 4 | 1 | 0 | 22.00 | 2.65 | | |
| 4 | 4 | 1 | 0 | 23.20 | 3.40 | 23.05 | 3.13 |
| 5 | 4 | 1 | 0 | 23.50 | 2.74 | | |
| 6 | 4 | 1 | 0 | 24.20 | 3.23 | | |
| 7 | 4 | 1 | 0 | 25.40 | 3.35 | | |
| 8 | 4 | 1 | 0 | 26.30 | 3.44 | | |
| 9 | 4 | 1 | 45 | 13.50 | 2.58 | | |
| 10 | 4 | 1 | 45 | 14.70 | 2.41 | | |
| 11 | 4 | 1 | 45 | 14.75 | 2.63 | | |
| 12 | 4 | 1 | 45 | 15.85 | 3.82 | 15.69 | 3.02 |
| 13 | 4 | 1 | 45 | 16.60 | 3.65 | | |
| 14 | 4 | 1 | 45 | 17.10 | 2.62 | | |
| 15 | 4 | 1 | 45 | 17.30 | 3.44 | | |
| ... | ... | ... | ... | ... | ... | ... | ... |
| 24 | 4 | 2 | 0 | 12.00 | 2.35 | 14.18 | 2.23 |
| 25 | 4 | 2 | 0 | 13.30 | 1.68 | | |
| ... | ... | ... | ... | ... | ... | | |
| 31 | 4 | 2 | 45 | 18.50 | 2.30 | 21.44 | 2.99 |
| 32 | 4 | 2 | 45 | 20.20 | 3.06 | | |
| ... | ... | ... | ... | ... | ... | | |
| 45 | 4 | 2 | 90 | 17.40 | 2.88 | 16.28 | 2.96 |
| 46 | 4 | 2 | 90 | 19.50 | 3.62 | | |

The other half was obtained by repeating the same values as for the first half, except for the angle of the tensile force which was modified by adding 180° (reversed tensile forces). The training was done repeatedly for a number of neurons of the intermediate layer that belonged to the interval [10,25]. The network kept had the smallest mean square error (MSE=1.43) and 15 neurons in the hidden layer.

*B. ANN Validation*

The validation of the network following the training was done using the data in Table II organized as follows:

- Inputs - columns 3 and 4;
- Actual outputs - columns 5 and 6;
- Simulated outputs - columns 7 and 8.

The mean square error calculated at validation was MSE=1.12.

**TABLE II**
**DATA USED IN THE VALIDATION PHASE OF THE ANN**

| No. | Number of layers | Inputs | | Actual outputs | | Simulated outputs | | Absolute errors | | Relative errors % | |
|---|---|---|---|---|---|---|---|---|---|---|---|
| | | Layout of the layers | Angle of tensile | $\sigma_M$ | $\varepsilon_M$ | $\sigma_{Msim}$ | $\varepsilon_{Msim}$ | $\Delta\sigma_M$ | $\Delta\varepsilon_M$ | $\Delta\sigma_M$ (%) | $\Delta\varepsilon_M$(%) |
| (1) | (2) | (3) | (4) | (5) | (6) | (7) | (8) | (9) | (10) | (11) | (12) |
| 1 | 4 | 1 | 0 | 22.10 | 2.82 | 23.05 | 3.13 | -0.95 | -0.31 | -4.30 | -10.90 |
| 2 | 4 | 1 | 0 | 24.60 | 3.38 | 23.05 | 3.13 | 1.55 | 0.25 | 6.30 | 7.47 |
| 3 | 4 | 1 | 45 | 14.30 | 3.20 | 15.73 | 2.80 | -1.43 | 0.40 | -10.03 | 12.42 |
| 4 | 4 | 1 | 45 | 16.23 | 3.80 | 15.73 | 2.80 | 0.50 | 1.00 | 3.06 | 26.25 |
| 5 | 4 | 1 | 45 | 17.90 | 3.45 | 15.73 | 2.80 | 2.17 | 0.65 | 12.10 | 18.77 |
| 6 | 4 | 1 | 90 | 14.95 | 2.66 | 14.97 | 2.77 | -0.02 | -0.11 | -0.16 | -4.15 |
| 7 | 4 | 1 | 90 | 16.80 | 2.83 | 14.97 | 2.77 | 1.83 | 0.06 | 10.87 | 2.11 |
| 8 | 4 | 2 | 0 | 13.10 | 2.90 | 14.18 | 2.23 | -1.08 | 0.67 | -8.23 | 23.03 |
| 9 | 4 | 2 | 0 | 14.40 | 2.40 | 14.18 | 2.23 | 0.22 | 0.17 | 1.54 | 6.99 |
| 10 | 4 | 2 | 0 | 17.00 | 2.63 | 14.18 | 2.23 | 2.82 | 0.40 | 16.60 | 15.13 |
| 11 | 4 | 2 | 45 | 21.40 | 2.84 | 21.42 | 3.04 | -0.02 | -0.20 | -0.09 | -7.12 |
| 12 | 4 | 2 | 45 | 22.30 | 3.41 | 21.42 | 3.04 | 0.88 | 0.37 | 3.95 | 10.79 |
| 13 | 4 | 2 | 90 | 14.30 | 2.45 | 16.30 | 2.83 | -2.00 | -0.38 | -13.98 | -15.36 |
| 14 | 4 | 2 | 90 | 16.50 | 2.64 | 16.30 | 2.83 | 0.20 | -0.19 | 1.22 | -7.06 |
| 15 | 4 | 2 | 90 | 18.10 | 3.10 | 16.30 | 2.83 | 1.80 | 0.27 | 9.95 | 8.83 |

The fit between the actual outputs (tensile strength $\sigma_M$ and elongation at break $\varepsilon_M$) and the simulated ones ($\sigma_{Msim}$ and $\varepsilon_{Msim}$) is presented in Figure 4.

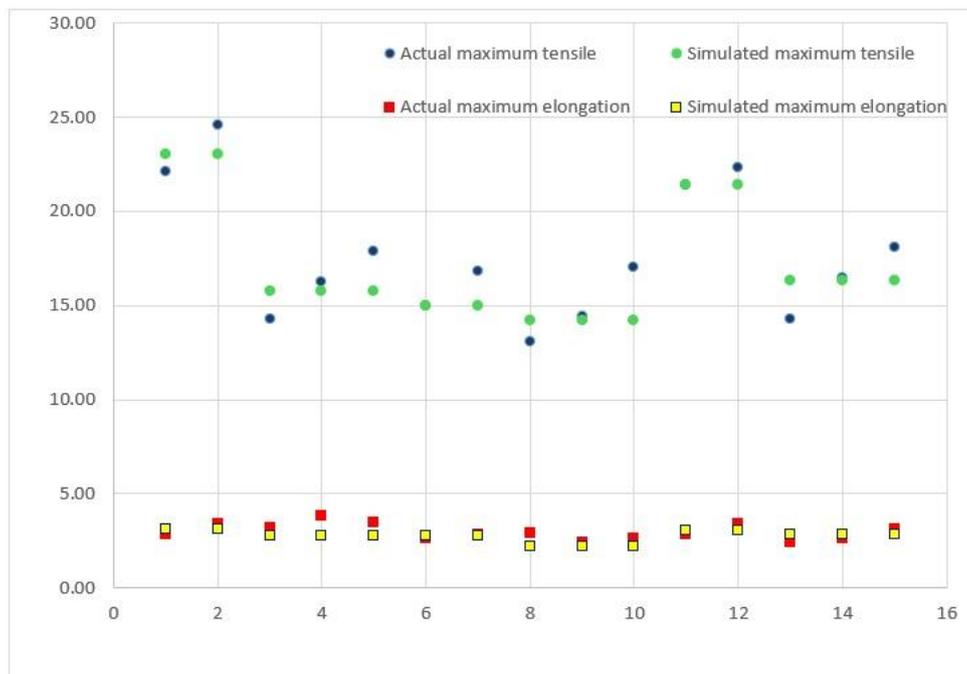

**Figure 4. Comparison between actual outputs vs. simulated outputs**

*C. ANN recall phase*

After the training and validation phase, the network, in its recall phase, was used to approximate the values of the tensile strength $\sigma_M$ and elongation at break $\varepsilon_M$ of the composite material for a set of 36 input values (pairs composed of the orientation of the layers and the angle of the tensile force).

The results are presented in Table III. The input data were chosen so as to form certain groups:
- Group 1: inputs that have the layers oriented the same (1 and 2), and the tensile forces opposed – pairs of rows (1,16), (3,17), (5,18), (30,35);
- Group 2: inputs that have the same values as some of the training data - (1,0), (1,45), (1,90), (2,0), (2,45), (2,90);
- Group 3: input pairs that have their layers oriented the same (1 or 2), and the traction forces acting on directions symmetrical to the direction of the training force – pairs of rows (1,4), (6,9), (11,14), (20,23), (25,28), (30,33);
- Group 4: the rest of the data.

TABLE III
RESULTS OBTAINED IN THE RECALL PHASE OF THE ANN

| No. | Inputs | | Outputs | | No. | Inputs | | Outputs | |
|---|---|---|---|---|---|---|---|---|---|
| | Layout of the layers | Angle of tensile | $\sigma_M$ | $\varepsilon_M$ | | Layout of the layers | Angle of tensile | $\sigma_M$ | $\varepsilon_M$ |
| 1 | 1 | -10 | 14.13 | 6.30 | 19 | 1 | 225 | 15.69 | 2.98 |
| 2 | 1 | 0 | 23.05 | 3.13 | 20 | 2 | -10 | 17.11 | 4.32 |
| 3 | 1 | 1 | 20.86 | 1.81 | 21 | 2 | 0 | 14.18 | 2.23 |
| 4 | 1 | 10 | 10.07 | -0.70 | 22 | 2 | 1 | 19.43 | 4.94 |
| 5 | 1 | 22 | 17.72 | 2.89 | 23 | 2 | 10 | 22.70 | 3.10 |
| 6 | 1 | 35 | 16.38 | 2.83 | 24 | 2 | 22 | 22.51 | 3.09 |
| 7 | 1 | 45 | 15.73 | 2.80 | 25 | 2 | 35 | 22.06 | 3.07 |
| 8 | 1 | 46 | 15.69 | 2.80 | 26 | 2 | 45 | 21.42 | 3.04 |
| 9 | 1 | 55 | 15.36 | 2.79 | 27 | 2 | 46 | 21.34 | 3.04 |
| 10 | 1 | 67 | 15.13 | 2.78 | 28 | 2 | 55 | 20.44 | 3.00 |
| 11 | 1 | 80 | 15.01 | 2.77 | 29 | 2 | 67 | 18.90 | 2.94 |
| 12 | 1 | 90 | 14.97 | 2.77 | 30 | 2 | 80 | 17.25 | 2.87 |
| 13 | 1 | 91 | 14.97 | 2.77 | 31 | 2 | 90 | 16.30 | 2.83 |
| 14 | 1 | 100 | 14.95 | 2.77 | 32 | 2 | 91 | 16.22 | 2.82 |
| 15 | 1 | 112 | 14.95 | 2.77 | 33 | 2 | 100 | 15.69 | 2.80 |
| 16 | 1 | 170 | 23.06 | 3.09 | 34 | 2 | 112 | 15.28 | 2.78 |
| 17 | 1 | 181 | 23.05 | 3.09 | 35 | 2 | 260 | 21.09 | 3.06 |
| 18 | 1 | 202 | 20.06 | 3.05 | 36 | 2 | 290 | 15.69 | 2.98 |

## IV. CONCLUSIONS

The following conclusions can be drawn after the modelling of the tensile strength $\sigma_M$ and the elongation at break $\varepsilon_M$:
1. Even though the network was trained on a relatively small set of examples (92 input-output pairs), the mean square error for the validation of the network is acceptable (MSE=1.12);
2. The degree of fit between the experimentally determined output variables and the ones simulated using the ANN (Figure 4, validation phase) is larger in the case of the elongation at break $\varepsilon_M$ than for the tensile strength $\sigma_M$. This is natural because the spread of the training values of $\sigma_M$ is larger than the one for the values of $\varepsilon_M$.
3. During the recall phase for the inputs that are part of group 2, the outputs represent average values of the outputs used for training. Example: the simulated output $(\sigma_M,\varepsilon_M)$ = (25.03,3.13) that corresponds to the input (1,0) (Table III) is equal to the average of the outputs in the training set that correspond to the same input (1,0) and which were experimentally measured (see Table I, rows 1 – 8).
4. During the recall phase, the outputs that correspond to the inputs that are part of group 1 should be close, but they aren't because of the lack of such examples in the training set.
5. The conclusion from the previous point also applies to group 3.